\documentclass[10pt,twocolumn,letterpaper]{article}

\usepackage{wacv}              

\usepackage{graphicx}
\usepackage{amsmath}
\usepackage{amssymb}
\usepackage{booktabs}
\usepackage{algorithm}
\usepackage{algpseudocode}
\usepackage[authoryear]{natbib}

\usepackage[pagebackref,breaklinks,colorlinks]{hyperref}

\usepackage[capitalize]{cleveref}
\crefname{section}{Sec.}{Secs.}
\Crefname{section}{Section}{Sections}
\Crefname{table}{Table}{Tables}
\crefname{table}{Tab.}{Tabs.}

\def\wacvPaperID{18} 
\def\confName{WACV}
\def\confYear{2025}

\begin{document}

\title{PQD: POST-TRAINING QUANTIZATION FOR EFFICIENT DIFFUSION MODELS}

\author{Jiaojiao Ye\\
University of Oxford1\\
\and
Zhen Wang\\
Zhejiang Lab2\\
\and
Linnan Jiang\\
Zhejiang Lab2\\
}
\maketitle

\begin{abstract}
   Diffusion models (DMs) have demonstrated remarkable achievements in synthesizing images of high fidelity and diversity. However, the extensive computational requirements and slow generative speed of diffusion models have limited their widespread adoption. In this paper, we propose a novel post-training quantization for diffusion models (PQD), which is a time-aware optimization framework for diffusion models based on post-training quantization. The proposed framework optimizes the inference process by conducting time-aware calibration. Experimental results show that our proposed method is able to directly quantize full-precision diffusion models into 8-bit or 4-bit models while maintaining comparable performance in a training-free manner, achieving a few FID change on ImageNet for unconditional image generation. Our approach demonstrates versatility and compatibility, and can also be applied to 512x512 text-guided image generation for the first time.
\end{abstract}

\section{Introduction}
\label{sec:intro}

Recently, diffusion models (DMs) have achieved phenomenal success in synthesizing high-fidelity and diverse images.
As a class of flexible generative models, like other generative models such as generative adversarial networks (GANs) \citep{goodfellow2020generative, Guo2019AGM}, variational autoencoder (VAE) \cite{kingma2013auto} 
, diffusion models demonstrate their power in various applications such as graph generation \cite{2020arXiv200300638N}, image-to-image translation \cite{2021arXiv210405358S} and molecular conformation generation \cite{xu2022geodiff}. In contrast to GANs and VAE, diffusion models avoid the issues of mode collapse and posterior collapse, leading to more stable training processes.

However, it is hard to adopt diffusion models widely because of high computational requirements and slow generative speed. The generation of high-quality outputs requires significant computational resources, particularly during each iterative step that involves transitioning from a noisy output to a less noisy one. For example, the execution of Stable Diffusion  \cite{rombach2021highresolution} necessitates 16GB of running memory and GPUs with over 10GB of VRAM, which is infeasible for most consumer-grade PCs and resource-constrained edge devices. An intriguing avenue of exploration is to bridge the sampling speed gap between DMs and GANs while preserving image quality. 

The slow inference in diffusion models can be attributed to two main factors: 1). The lengthy iteration in the denoising process 2). The complex network required for estimating the noise in each denoising iteration. Previous works \cite{song2019generative, song2020denoising, chen2020wavegrad} have attempted to address the computational challenges of diffusion models by shortening the denoising process, namely, less iteration in the denoising process. However, this approach can lead to a loss of image quality. 

Moreover, to accelerate diffusion models effectively, a training-free methodology is essential. This is because acquiring large training datasets for diffusion models is often challenging due to privacy and commercial concerns. Furthermore, the training process itself is highly resource-intensive and time-consuming. For example, training a class-conditional Latent Diffusion Model (LDM) on the ImageNet dataset requires 35 V100 GPU days \cite{rombach2021highresolution}. Therefore, we require network acceleration techniques that do not rely on training.

To address the above-mentioned issues, this research will explore optimization for diffusion models with a focus on enhancing the inference process for various image generation tasks. 
The contribution of this work can be threefold: 

\begin{enumerate}
    \item To optimize inference of the denoising diffusion models, we propose a time-aware post-training quantization on diffusion models (PQD), which includes a calibration dataset selection algorithm across multiple denoising steps.
    \item We extend our quantization method to latent space diffusion models and allow for high-quality text-guided image synthesis.
    \item We evaluate our method on unconditional image generation and text-to-image tasks and then compare it with existing PTQ methods on diffusion models. We achieve state-of-the-art results on 8-bits models for 512x512 high-resolution text-guided image generation. The results demonstrate that our method achieves better image quality than the existing methods with higher compression.
\end{enumerate}

\section{Related work}
\label{sec:relation}

\textbf{Sampling Acceleration for Diffusion Models}


To accelerate the inference of diffusion models, the mainstream approaches focus on shortening the sampling path, where four advanced techniques are used to enhance sampling speed: distillations \cite{salimans2022progressive, berthelot2023tract, zhang2023destseg, song2023consistency}, training schedule optimization \cite{hoogeboom2022blurring, kong2021fast, daras2022soft, nichol2021improved}, training-free acceleration \cite{song2019generative, song2020denoising, karras2022elucidating, bao2022analytic}, and integration of diffusion models with faster generative models \cite{lyu2022accelerating, pandey2022diffusevae, xiao2021tackling}. For example, Song \etal \cite{song2019generative} formulates the diffusion models as an ordinary differential equation (ODE) and improves sampling efficiency by using a faster ODE solver. One of the earliest works DDIM \cite{song2020denoising} accelerates diffusion models sampling by adopting implicit phases in the denoising process.
In summary, all of these methods are identified by finding effective sampling trajectories. 

\textbf{Post-training Quantization}

Quantization is a model compression technique that can significantly reduce the size and computational cost of the network, making it more efficient to deploy and use. 
There are two main categories of quantization algorithms: quantization-aware training (QAT) \cite{louizos2018relaxed, zhuang2018towards, gong2019differentiable} and post-training quantization (PTQ) \cite{levkovitch2022zero, wei2022qdrop, li2022patch, DBLP:conf/iconip/ChenYCWFCGH23}. QAT involves simulating quantization during the training process to maintain high performance at reduced precision levels. However, this method demands significant time, computational resources, and access to the original dataset. On the other hand, post-training quantization PTQ eliminates the need for fine-tuning and requires only a small quantity of unlabeled data for calibration.
For example, BRECQ \cite{Li2021BRECQPT} introduces fisher information into the objective, and optimizes layers within a single residual block jointly using a small subset of calibration data from the training dataset. 
Most of existing works have successfully applied quantization to convolutional network and its successors, and transformer-based architectures.

However, applying PTQ to diffusion models is challenging due to its multi-step inference, which is analyzed in-depth in \cref{ssec:observations}. To the best of our knowledge, PTQ4DM \cite{shang2022post} is the first work to accelerate generation from the perspective of compressing the noise estimation. They proposed a PTQ method for diffusion models that can directly quantize full-precision models into 8-bit models while maintaining or even improving their performance. Additionally, PTQD \cite{he2024ptqd} analyzed systematically the quantization effect on diffusion models and establish a unified framework for accurate post-training diffusion quantization. However, experiments in those researches were limited to small datasets and low resolution. Their methods may fail for high-resolution image generation and not support for text-to-image generation without conditional features. 

In this work, we extend the PTQ4DM method of \cite{shang2022post} to high-resolution image generation and propose an algorithm for latent-space image generation. Our experiments show that our method can significantly achieve comparative great quality in the generation of high-resolution images (512x512). 

\section{Methodology}
\label{sec:method}

In this section, we will begin by revisiting the key concepts outlined in \cref{sec:sd} and \cref{ssec:observations}. Following that, we will delve into the challenges posed to existing (post-training quantization) PTQ calibration methods when applied to diffusion models in \cref{ssec:observations}. Subsequently, we will introduce our novel approach for performing post-training quantization on diffusion models in \cref{ssec:PQD}.

\subsection{Diffusion Models}
\label{sec:sd}

Diffusion models (DMs) \cite{Song2019GenerativeMB} aim to generate images by utilizing the Markov chain, which consists of two main processes. The forward process involves the gradual addition of isotropic noise with a variance schedule denoted by $\beta_{1}, \ldots, \beta_{T} \in (0, 1)$, where $\beta_{1}, \ldots, \beta_{T} \in (0, 1) $. This process generates a sequence of noise variables, denoted as $x_1, \ldots, x_T$. On the other hand, the reverse process, also referred to as the denoising process, gradually samples an image from a Gaussian noise distribution by following the conditional distribution $q(x_{t-1} \mid x_{t})$, defined in \cref{eq:mc}. 
 
\begin{equation}
q(x_{t-1} \mid x_{t}) = \mathcal{N}(x_t; \sqrt{1 - \beta_{t} }x_{t-1},\beta_{t} \mathbf{I} ) .
\label{eq:mc}
\end{equation}

Thus, we can express $x_t$ as a form of linear variable $x_0$ and noise:

\begin{equation}
x_t = \sqrt{\alpha_t} x_0 + \sqrt{1- \alpha_t} \epsilon, 
\end{equation}

where $\epsilon \sim  \mathcal{N}(0,\,\mathbf{I}), \alpha_{t} = 1 - \beta_{t}$. From a trained model, $x_0$ is sampled by first sampling $x_T$ from the prior $p_{\theta}(x_{T})$, and then sampling $x_{t-1}$ from the denoising processes iteratively.

\subsection{Post-training Quantization}
\label{sec:PTQ}
Post-training quantization (PTQ) \cite{Nagel2019DataFreeQT} techniques enable quantization without necessitating retraining, making them particularly suitable for scenarios with limited data and user-friendly applications. 
In this context, the transformation of a tensor into a quantized tensor is orchestrated by the use of quantization parameters, namely the scaling factor $s$ and the zero point $z$. The quantization-dequantization process can be described as follows:

\begin{equation}
    \mathbf{X}_{sim} = s \left( clamp\left( \frac{\mathbf{X}}{s}, p_{min}, p_{max} \right) + z\right),
    \label{eq1}
\end{equation}

where $p_{min}$, $p_{max}$ signify the threshold defined by bitwidth, $\mathbf{X}_{sim}$ is the de-quantized tensor. The determination of suitable quantization parameters is based on the minimization of the calibration error. This error is quantified by the calibration loss, where the L2 distance can serve as viable metrics. 


In general, PTQ quantizes a network through three steps: (i) Identify which operations within the network should be quantized, leaving the remaining operations in full precision. (ii) Gather calibration dataset. To prevent overfitting of quantization parameters to the calibration samples, their distribution should closely match that of the real data. (iii) Employ an appropriate method to determine quantization parameters for weight and activation tensors.



\begin{figure}[ht]
    \centering
    \includegraphics*[width=0.45\textwidth]{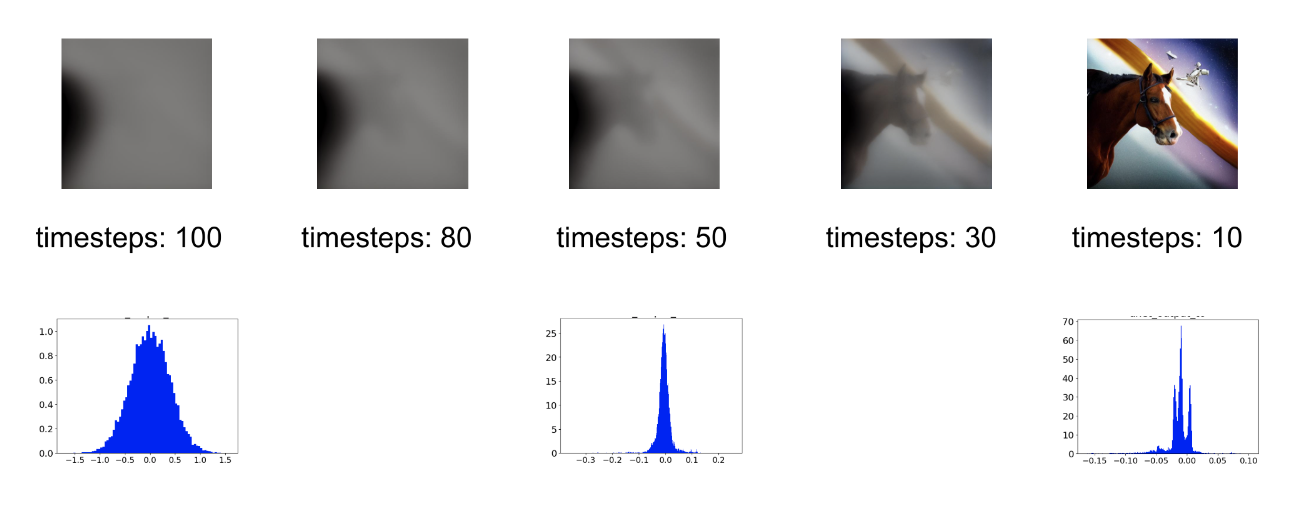}
    \caption{The activation of the output layer varies during the denoising process. }
    \label{fig:property1}
\end{figure}

\begin{figure*}[ht!]
    \centering
    \includegraphics[width=0.75\textwidth]{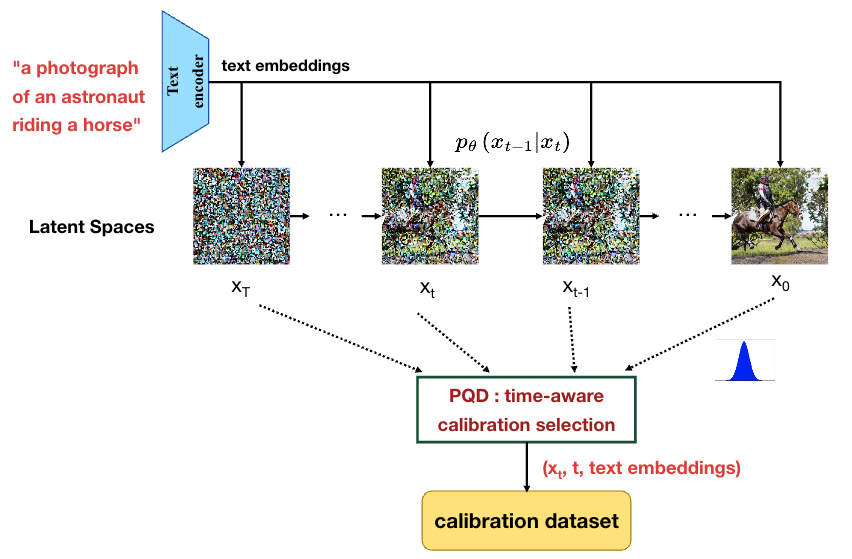}
    \caption{PQD construct calibration dataset with time step distributed over denoising process. Our method generate inputs that are accurate reflections of data seen during the production in a data-free manner.}
    \label{fig:pqd}
\end{figure*}

\subsection{Exploration of PTQ on DMs}
\label{ssec:observations}


In order to understand the change in the output distribution of diffusion models, we investigate the activation distribution with respect to the time step. If the distribution changes with respect to the time step, it could pose a challenge for the adoption of previous PTQ calibration methodologies, as they are proposed for temporally invariant calibration. Previous studies \cite{shang2022post} 
have also reported the dynamic property of activation in diffusion models and attempted to address it by sampling the calibration dataset across all time frames.

\cref{fig:property1} illustrates the activation of noise estimation network, a U-Net \cite{ronneberger2015u} over the denoising process. We could observe that the activation distribution of diffusion models changes over the denoising process, which poses intricate challenges to the quantization process. This is because that quantization is applied for each operation, which means that we only have a fixed parameter for one operation. Therefore, it is necessary to ensure the activation of diffusion models could represent the main features during the denoising process.




\subsection{Post-training Quantization for DMs (PQD)}
\label{ssec:PQD}

Based on our previous observations, we proposed a time-aware post-training quantization for diffusion models (PQD). The process is to first select a calibration dataset by our calibration algorithm, and then apply QDrop \cite{wei2022qdrop} on pretrained diffusion models to quantize the model.


As diffusion models are quite different from classic CNNs and ViT, it is necessary to design a novel and effective calibration dataset collection method in this multi-time-step scenario. To recover the performance of the quantized diffusion models, we need to select calibration data in a way that closely mirrors the true output distribution of different time steps. As \cref{fig:pqd} shows, we design a DM-specific collection strategy for dynamic activation iteration in our PQD framework. Specifically, we sample calibration following a Gaussian Distribution. The benefit of this normally distributed time-step is that we can sample activation over the denoising process as our calibration dataset. By choosing different $\mu$ and $\sigma$, this strategy has the flexibility to let activation be more similar to synthetic image or noise.


\subsection{Extension to High-Resolution Image Generation}
High-resolution image generation is challenging because the computational cost increases exponentially with the size of the image. To enable efficient high-resolution image generation, we modified our framework for high-resolution image synthesis. Specifically, we extracted the latent feature and applied quantization in the latent feature instead of input features. This allows efficient low-bit inference on 512x512 text-guided image generation.


For text-guided image generation with Stable Diffusion, we need to also include text conditioning in the calibration dataset. This specific calibration dataset creation process is described by \cref{alg:NDTC4sd}. Assume we need to collect a calibration dataset of size N, the time step selection follows distribution $\mathcal{N}(\mu,\,\sigma)$. For each prompt we add a pair of data with both a conditional feature $c_t$ and an unconditional feature $uc_t$ derived from the prompt.

\begin{algorithm}
    \caption{PQD calibration for Text-Guided Image Generation} \label{alg:NDTC4sd}
    \begin{algorithmic}[1]
    \Statex \textbf{Input:} Empty calibration dataset $\mathcal{D}$
    \Statex \textbf{Require:} The size of the calibration dataset $N$, Number of denoising steps $T$, normal distribution mean $\mu$, standard deviation $\sigma$
    \For{\texttt{$i = 1$ to $N$}}
        \State  Sample $t_i$ from a skew normal distribution $\mathcal{N}(\mu,\,\sigma)$;
        \State Round down $t_i$ into an integer;
        \State Clamp $t_{i}$ between $[0, T]$;
        \State Generate a gaussian noise $x_T$ as initialization;        
        \For{$t = T$, $\cdots$ , $t_i$ time steps}
        \State Sample intermediate variables from full-precision  noising estimation model;
        \EndFor
        \State Sample intermediate inputs $\left( x_{t_{i}}, c_{t_{i}}, t_{t_{i}} \right)$, $\left( x_{t_{i}}, uc_{t_{i}}, t_{t_{i}} \right)$ and add them to calibration dataset $\mathcal{D}$;
        
      \EndFor
    \State Apply QDrops with full-precision diffusion model and $\mathcal{D}$.
    \Statex \textbf{Output:} Quantized model.
    \end{algorithmic}
\end{algorithm}

\label{ssec:high-res}

\section{Experiments}
\label{sec:experiment}

\begin{figure*}[ht!]
    \centering
    \includegraphics*[width=0.9\textwidth]{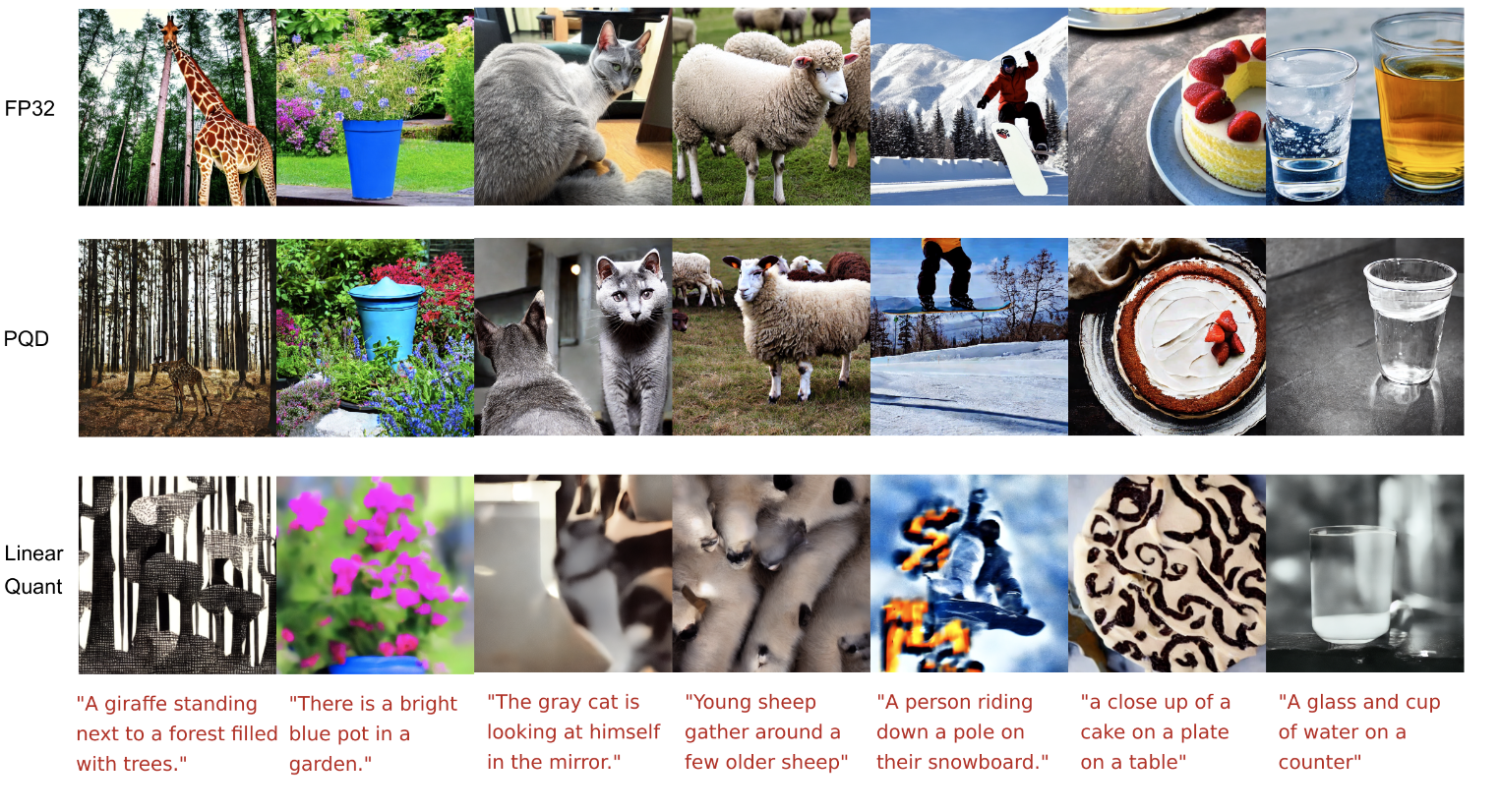}
    \caption{Text-guided generated samples with 512x512 resolution by Stable Diffusion model. \textbf{Upper} Samples generated using the full-precision model. \textbf{Middle} Samples generated by our 8-bit quantized model. \textbf{Bottom} Samples generated by 8-bit Linear Quantization model.}
    \label{fig:generated samples}
\end{figure*}


\subsection{Experiment Setup}
\label{ssec: setting }
We conduct a series of image synthesis experiments using Denoising Diffusion Probabilistic Models (DDPM) \cite{ho2020denoising} pretrained on downsampled ImageNet \cite{imagetnet} for unconditional generation and Stable Diffusion \cite{rombach2021highresolution} pretrained on subsets of 512×512 LAION-5B \cite{schuhmann2022laion} for text-guided image generation. We experiment on two standard benchmarks: ImageNet and MS-COCO \cite{lin2014microsoft}. For comparative reference, we establish PTQ4DM \cite{shang2022post}.



To obtain a comparable test field, we maintain the computational resources to a single NVIDIA A100 GPU for all experiments in this section. To make a trade-off between optimization and image quality, we will evaluate the experiment on Inception Score (IS) \cite{salimans2016improved}, Fréchet Inception Distance (FID) \cite{heusel2017gans} to measure image fidelity. 
Bops is calculated for one denoising step without considering the decoder compute cost for latent diffusion.

\subsection{Unconditional Generation}
\label{ssec:ug}
\begin{table}
\centering
\label{table:imagenet}
\resizebox{\columnwidth}{!}{%
\begin{tabular}{lllllll}
\hline 
 Method & Bits(W/A) & Size (Mb) & GBops &  FID$\downarrow$ &  IS $\uparrow$ \\
 \hline
 Full Precision & W32A32 & 143.2 & 6597 & 21.63 & 14.88 
\\
 \hline
  PQD & W4A32 & 17.9 & 1147 & 69.14 & 4.40  \\ 
 \hline
  PTQ4DM & W8A8 & 35.8 & 798 & 23.96 & 15.88  \\
  \hline
  PQD & W8A8 & 35.8 & 798 &  \textbf{22.29}  & \textbf{16.21 }\\ 
 \hline
  PQD & W4A8 & \textbf{17.9} & \textbf{399} & 75.32 & 4.48 \\ 
  \hline 
\end{tabular}
}
\caption{Quantization results for unconditional image generation with DDIM 250 steps on ImageNet (64×64). The numbers inside the PTQ4DM parentheses refer to \cite{shang2022post} results with INT8 attention act-to-act matmuls. W8A8 represents 8 bits weight quantization and 8 bits activation quantization.}

\end{table}

In an unconditional generation experiment, we use the pre-trained DDIM sampler with 250 denoising time steps for ImageNet synthesis. The detailed parameters of calibration algorithm are as the following: $N=5120$, $\mu = 0.4$, $\sigma =0.4$. These parameters are attained by tuning, and theoretically it would benefit to collect calibration dataset relatively similar to $x_0$, far away from $x_T$.


The results are reported in \cref{table:imagenet}.
The experiments show that our proposed PQD method significantly preserves the image generation quality and outperforms the baseline. Although 8-bit weight quantization has almost no performance loss compared to FP32 for both PTQ4DM and our approach, under 4-bit weight quantization, our methods still preserve most of the perceptual quality with few increase in FID and imperceptible distortions in produced samples. 


\subsection{Text-to-image Generation}
\label{ssec:txt2img}
Secondly, we evaluate the quantization of Stable Diffusion for high-resolution text-to-image generation on MS-COCO. 
We sample prompts from the MS-COCO dataset to generate a calibration dataset with text conditions. Note that we choose 5120 batches from MS-COCO 2017 Val as text in calibration data by recovering the distribution of the training dataset. The detailed parameters of calibration algorithm are as the following: $N=5120$, $\mu = 0.4$, $\sigma =0.4$.


The qualitative results of high-resolution generation quantization are presented in \cref{fig:generated samples}. In some cases, directly applying naive Linear Quantization degrades the appearance of forests, people, cats \etc. Compared to Linear Quantization, our PQD provides more realistic and higher-quality images with more details and better demonstration of the semantic information.  




\section{Conclusion}
\label{sec:conclusion}

This paper introduces PQD, a novel training-free, time-aware post-training quantization framework for diffusion models, which optimizes the inference process through time-aware calibration across multiple denoising steps. By directly quantizing full-precision diffusion models into 8-bit or 4-bit representations, PQD maintains performance levels comparable to the original models without requiring additional training. Our method achieves significant improvements in the high-resolution performance for text-to-image generation task under 8-bit quantization. 
Furthermore, the framework's versatility and compatibility are evident in its successful application to large-scale image generation and 512x512 text-guided image synthesis, thereby expanding the potential use cases of diffusion models in various domains.




{\small
\bibliographystyle{plainnat}
\bibliography{main}

\begin{thebibliography}{42}
\providecommand{\natexlab}[1]{#1}
\providecommand{\url}[1]{\texttt{#1}}
\expandafter\ifx\csname urlstyle\endcsname\relax
  \providecommand{\doi}[1]{doi: #1}\else
  \providecommand{\doi}{doi: \begingroup \urlstyle{rm}\Url}\fi

\bibitem[Bao et~al.(2022)Bao, Li, Zhu, and Zhang]{bao2022analytic}
Fan Bao, Chongxuan Li, Jun Zhu, and Bo~Zhang.
\newblock Analytic-dpm: an analytic estimate of the optimal reverse variance in diffusion probabilistic models.
\newblock \emph{arXiv preprint arXiv:2201.06503}, 2022.

\bibitem[Berthelot et~al.(2023)Berthelot, Autef, Lin, Yap, Zhai, Hu, Zheng, Talbott, and Gu]{berthelot2023tract}
David Berthelot, Arnaud Autef, Jierui Lin, Dian~Ang Yap, Shuangfei Zhai, Siyuan Hu, Daniel Zheng, Walter Talbott, and Eric Gu.
\newblock Tract: Denoising diffusion models with transitive closure time-distillation.
\newblock \emph{arXiv preprint arXiv:2303.04248}, 2023.

\bibitem[Chen et~al.(2020)Chen, Zhang, Zen, Weiss, Norouzi, and Chan]{chen2020wavegrad}
Nanxin Chen, Yu~Zhang, Heiga Zen, Ron~J Weiss, Mohammad Norouzi, and William Chan.
\newblock Wavegrad: Estimating gradients for waveform generation.
\newblock \emph{arXiv preprint arXiv:2009.00713}, 2020.

\bibitem[Chen et~al.(2023)Chen, Yan, Cheng, Wang, Fu, Chen, Guan, and He]{DBLP:conf/iconip/ChenYCWFCGH23}
Xinrui Chen, Renao Yan, Junru Cheng, Yizhi Wang, Yuqiu Fu, Yi~Chen, Tian Guan, and Yonghong He.
\newblock Adeq: Adaptive diversity enhancement for zero-shot quantization.
\newblock In \emph{ICONIP (1)}, pages 53--64, 2023.
\newblock URL \url{https://doi.org/10.1007/978-981-99-8079-6_5}.

\bibitem[Daras et~al.(2022)Daras, Delbracio, Talebi, Dimakis, and Milanfar]{daras2022soft}
Giannis Daras, Mauricio Delbracio, Hossein Talebi, Alexandros~G Dimakis, and Peyman Milanfar.
\newblock Soft diffusion: Score matching for general corruptions.
\newblock \emph{arXiv preprint arXiv:2209.05442}, 2022.

\bibitem[Deng et~al.(2009)Deng, Dong, Socher, Li, Li, and Fei-Fei]{imagetnet}
Jia Deng, Wei Dong, Richard Socher, Li-Jia Li, Kai Li, and Li~Fei-Fei.
\newblock Imagenet: A large-scale hierarchical image database.
\newblock In \emph{2009 IEEE Conference on Computer Vision and Pattern Recognition}, pages 248--255, 2009.
\newblock \doi{10.1109/CVPR.2009.5206848}.

\bibitem[Gong et~al.(2019)Gong, Liu, Jiang, Li, Hu, Lin, Yu, and Yan]{gong2019differentiable}
Ruihao Gong, Xianglong Liu, Shenghu Jiang, Tianxiang Li, Peng Hu, Jiazhen Lin, Fengwei Yu, and Junjie Yan.
\newblock Differentiable soft quantization: Bridging full-precision and low-bit neural networks.
\newblock In \emph{Proceedings of the IEEE/CVF international conference on computer vision}, pages 4852--4861, 2019.

\bibitem[Goodfellow et~al.(2020)Goodfellow, Pouget-Abadie, Mirza, Xu, Warde-Farley, Ozair, Courville, and Bengio]{goodfellow2020generative}
Ian Goodfellow, Jean Pouget-Abadie, Mehdi Mirza, Bing Xu, David Warde-Farley, Sherjil Ozair, Aaron Courville, and Yoshua Bengio.
\newblock Generative adversarial networks.
\newblock \emph{Communications of the ACM}, 63\penalty0 (11):\penalty0 139--144, 2020.

\bibitem[Guo et~al.(2019)Guo, Matthes, Ye, and Shen]{Guo2019AGM}
Mingpan Guo, Stefan Matthes, Jiaojiao Ye, and Hao Shen.
\newblock A generative map for image-based camera localization.
\newblock \emph{ArXiv}, abs/1902.11124, 2019.
\newblock URL \url{https://api.semanticscholar.org/CorpusID:67856346}.

\bibitem[He et~al.(2024)He, Liu, Liu, Wu, Zhou, and Zhuang]{he2024ptqd}
Yefei He, Luping Liu, Jing Liu, Weijia Wu, Hong Zhou, and Bohan Zhuang.
\newblock Ptqd: Accurate post-training quantization for diffusion models.
\newblock \emph{Advances in Neural Information Processing Systems}, 36, 2024.

\bibitem[Heusel et~al.(2017)Heusel, Ramsauer, Unterthiner, Nessler, and Hochreiter]{heusel2017gans}
Martin Heusel, Hubert Ramsauer, Thomas Unterthiner, Bernhard Nessler, and Sepp Hochreiter.
\newblock Gans trained by a two time-scale update rule converge to a local nash equilibrium.
\newblock \emph{Advances in neural information processing systems}, 30, 2017.

\bibitem[Ho et~al.(2020)Ho, Jain, and Abbeel]{ho2020denoising}
Jonathan Ho, Ajay Jain, and Pieter Abbeel.
\newblock Denoising diffusion probabilistic models.
\newblock \emph{Advances in neural information processing systems}, 33:\penalty0 6840--6851, 2020.

\bibitem[Hoogeboom and Salimans(2022)]{hoogeboom2022blurring}
Emiel Hoogeboom and Tim Salimans.
\newblock Blurring diffusion models.
\newblock \emph{arXiv preprint arXiv:2209.05557}, 2022.

\bibitem[Karras et~al.(2022)Karras, Aittala, Aila, and Laine]{karras2022elucidating}
Tero Karras, Miika Aittala, Timo Aila, and Samuli Laine.
\newblock Elucidating the design space of diffusion-based generative models.
\newblock \emph{Advances in neural information processing systems}, 35:\penalty0 26565--26577, 2022.

\bibitem[Kingma and Welling(2013)]{kingma2013auto}
Diederik~P. Kingma and Max Welling.
\newblock Auto-encoding variational bayes.
\newblock \emph{CoRR}, abs/1312.6114, 2013.
\newblock URL \url{https://api.semanticscholar.org/CorpusID:216078090}.

\bibitem[Kong and Ping(2021)]{kong2021fast}
Zhifeng Kong and Wei Ping.
\newblock On fast sampling of diffusion probabilistic models.
\newblock \emph{arXiv preprint arXiv:2106.00132}, 2021.

\bibitem[Levkovitch et~al.(2022)Levkovitch, Nachmani, and Wolf]{levkovitch2022zero}
Alon Levkovitch, Eliya Nachmani, and Lior Wolf.
\newblock Zero-shot voice conditioning for denoising diffusion tts models.
\newblock \emph{arXiv preprint arXiv:2206.02246}, 2022.

\bibitem[Li et~al.(2021)Li, Gong, Tan, Yang, Hu, Zhang, Yu, Wang, and Gu]{Li2021BRECQPT}
Yuhang Li, Ruihao Gong, Xu~Tan, Yang Yang, Peng Hu, Qi~Zhang, Fengwei Yu, Wei Wang, and Shi Gu.
\newblock Brecq: Pushing the limit of post-training quantization by block reconstruction.
\newblock \emph{ArXiv}, abs/2102.05426, 2021.

\bibitem[Li et~al.(2022)Li, Ma, Chen, Xiao, and Gu]{li2022patch}
Zhikai Li, Liping Ma, Mengjuan Chen, Junrui Xiao, and Qingyi Gu.
\newblock Patch similarity aware data-free quantization for vision transformers.
\newblock In \emph{European conference on computer vision}, pages 154--170. Springer, 2022.

\bibitem[Lin et~al.(2014)Lin, Maire, Belongie, Hays, Perona, Ramanan, Doll{\'a}r, and Zitnick]{lin2014microsoft}
Tsung-Yi Lin, Michael Maire, Serge Belongie, James Hays, Pietro Perona, Deva Ramanan, Piotr Doll{\'a}r, and C~Lawrence Zitnick.
\newblock Microsoft coco: Common objects in context.
\newblock In \emph{Computer Vision--ECCV 2014: 13th European Conference, Zurich, Switzerland, September 6-12, 2014, Proceedings, Part V 13}, pages 740--755. Springer, 2014.

\bibitem[Louizos et~al.(2018)Louizos, Reisser, Blankevoort, Gavves, and Welling]{louizos2018relaxed}
Christos Louizos, Matthias Reisser, Tijmen Blankevoort, Efstratios Gavves, and Max Welling.
\newblock Relaxed quantization for discretized neural networks.
\newblock \emph{arXiv preprint arXiv:1810.01875}, 2018.

\bibitem[Lyu et~al.(2022)Lyu, Xu, Yang, Lin, and Dai]{lyu2022accelerating}
Zhaoyang Lyu, Xudong Xu, Ceyuan Yang, Dahua Lin, and Bo~Dai.
\newblock Accelerating diffusion models via early stop of the diffusion process.
\newblock \emph{arXiv preprint arXiv:2205.12524}, 2022.

\bibitem[Nagel et~al.(2019)Nagel, van Baalen, Blankevoort, and Welling]{Nagel2019DataFreeQT}
Markus Nagel, Mart van Baalen, Tijmen Blankevoort, and Max Welling.
\newblock Data-free quantization through weight equalization and bias correction.
\newblock \emph{2019 IEEE/CVF International Conference on Computer Vision (ICCV)}, pages 1325--1334, 2019.
\newblock URL \url{https://api.semanticscholar.org/CorpusID:184487878}.

\bibitem[Nichol and Dhariwal(2021)]{nichol2021improved}
Alexander~Quinn Nichol and Prafulla Dhariwal.
\newblock Improved denoising diffusion probabilistic models.
\newblock In \emph{International conference on machine learning}, pages 8162--8171. PMLR, 2021.

\bibitem[Niu et~al.(2020)Niu, Song, Song, Zhao, Grover, and Ermon]{2020arXiv200300638N}
Chenhao Niu, Yang Song, Jiaming Song, Shengjia Zhao, Aditya Grover, and Stefano Ermon.
\newblock Permutation invariant graph generation via score-based generative modeling.
\newblock In \emph{International Conference on Artificial Intelligence and Statistics}, 2020.
\newblock URL \url{https://api.semanticscholar.org/CorpusID:211677799}.

\bibitem[Pandey et~al.(2022)Pandey, Mukherjee, Rai, and Kumar]{pandey2022diffusevae}
Kushagra Pandey, Avideep Mukherjee, Piyush Rai, and Abhishek Kumar.
\newblock Diffusevae: Efficient, controllable and high-fidelity generation from low-dimensional latents.
\newblock \emph{arXiv preprint arXiv:2201.00308}, 2022.

\bibitem[Rombach et~al.(2021)Rombach, Blattmann, Lorenz, Esser, and Ommer]{rombach2021highresolution}
Robin Rombach, Andreas Blattmann, Dominik Lorenz, Patrick Esser, and Björn Ommer.
\newblock High-resolution image synthesis with latent diffusion models, 2021.

\bibitem[Ronneberger et~al.(2015)Ronneberger, Fischer, and Brox]{ronneberger2015u}
Olaf Ronneberger, Philipp Fischer, and Thomas Brox.
\newblock U-net: Convolutional networks for biomedical image segmentation.
\newblock In \emph{Medical image computing and computer-assisted intervention--MICCAI 2015: 18th international conference, Munich, Germany, October 5-9, 2015, proceedings, part III 18}, pages 234--241. Springer, 2015.

\bibitem[Salimans and Ho(2022)]{salimans2022progressive}
Tim Salimans and Jonathan Ho.
\newblock Progressive distillation for fast sampling of diffusion models.
\newblock \emph{arXiv preprint arXiv:2202.00512}, 2022.

\bibitem[Salimans et~al.(2016)Salimans, Goodfellow, Zaremba, Cheung, Radford, and Chen]{salimans2016improved}
Tim Salimans, Ian Goodfellow, Wojciech Zaremba, Vicki Cheung, Alec Radford, and Xi~Chen.
\newblock Improved techniques for training gans.
\newblock \emph{Advances in neural information processing systems}, 29, 2016.

\bibitem[Sasaki et~al.(2021)Sasaki, Willcocks, and Breckon]{2021arXiv210405358S}
Hiroshi Sasaki, Chris~G. Willcocks, and T.~Breckon.
\newblock Unit-ddpm: Unpaired image translation with denoising diffusion probabilistic models.
\newblock \emph{ArXiv}, abs/2104.05358, 2021.
\newblock URL \url{https://api.semanticscholar.org/CorpusID:233210328}.

\bibitem[Schuhmann et~al.(2022)Schuhmann, Beaumont, Vencu, Gordon, Wightman, Cherti, Coombes, Katta, Mullis, Wortsman, et~al.]{schuhmann2022laion}
Christoph Schuhmann, Romain Beaumont, Richard Vencu, Cade Gordon, Ross Wightman, Mehdi Cherti, Theo Coombes, Aarush Katta, Clayton Mullis, Mitchell Wortsman, et~al.
\newblock Laion-5b: An open large-scale dataset for training next generation image-text models.
\newblock \emph{Advances in Neural Information Processing Systems}, 35:\penalty0 25278--25294, 2022.

\bibitem[Shang et~al.(2022)Shang, Yuan, Xie, Wu, and Yan]{shang2022post}
Yuzhang Shang, Zhihang Yuan, Bin Xie, Bingzhe Wu, and Yan Yan.
\newblock Post-training quantization on diffusion models.
\newblock \emph{arXiv preprint arXiv:2211.15736}, 2022.

\bibitem[Song et~al.(2020)Song, Meng, and Ermon]{song2020denoising}
Jiaming Song, Chenlin Meng, and Stefano Ermon.
\newblock Denoising diffusion implicit models.
\newblock \emph{arXiv preprint arXiv:2010.02502}, 2020.

\bibitem[Song and Ermon(2019{\natexlab{a}})]{Song2019GenerativeMB}
Yang Song and Stefano Ermon.
\newblock Generative modeling by estimating gradients of the data distribution.
\newblock In \emph{Neural Information Processing Systems}, 2019{\natexlab{a}}.
\newblock URL \url{https://api.semanticscholar.org/CorpusID:196470871}.

\bibitem[Song and Ermon(2019{\natexlab{b}})]{song2019generative}
Yang Song and Stefano Ermon.
\newblock Generative modeling by estimating gradients of the data distribution.
\newblock \emph{Advances in neural information processing systems}, 32, 2019{\natexlab{b}}.

\bibitem[Song et~al.(2023)Song, Dhariwal, Chen, and Sutskever]{song2023consistency}
Yang Song, Prafulla Dhariwal, Mark Chen, and Ilya Sutskever.
\newblock Consistency models.
\newblock \emph{arXiv preprint arXiv:2303.01469}, 2023.

\bibitem[Wei et~al.(2022)Wei, Gong, Li, Liu, and Yu]{wei2022qdrop}
Xiuying Wei, Ruihao Gong, Yuhang Li, Xianglong Liu, and Fengwei Yu.
\newblock Qdrop: Randomly dropping quantization for extremely low-bit post-training quantization.
\newblock \emph{arXiv preprint arXiv:2203.05740}, 2022.

\bibitem[Xiao et~al.(2021)Xiao, Kreis, and Vahdat]{xiao2021tackling}
Zhisheng Xiao, Karsten Kreis, and Arash Vahdat.
\newblock Tackling the generative learning trilemma with denoising diffusion gans.
\newblock \emph{arXiv preprint arXiv:2112.07804}, 2021.

\bibitem[Xu et~al.(2022)Xu, Yu, Song, Shi, Ermon, and Tang]{xu2022geodiff}
Minkai Xu, Lantao Yu, Yang Song, Chence Shi, Stefano Ermon, and Jian Tang.
\newblock Geodiff: A geometric diffusion model for molecular conformation generation.
\newblock In \emph{International Conference on Learning Representations}, 2022.
\newblock URL \url{https://openreview.net/forum?id=PzcvxEMzvQC}.

\bibitem[Zhang et~al.(2023)Zhang, Li, Li, Huang, Shan, and Chen]{zhang2023destseg}
Xuan Zhang, Shiyu Li, Xi~Li, Ping Huang, Jiulong Shan, and Ting Chen.
\newblock Destseg: Segmentation guided denoising student-teacher for anomaly detection.
\newblock In \emph{Proceedings of the IEEE/CVF Conference on Computer Vision and Pattern Recognition}, pages 3914--3923, 2023.

\bibitem[Zhuang et~al.(2018)Zhuang, Shen, Tan, Liu, and Reid]{zhuang2018towards}
Bohan Zhuang, Chunhua Shen, Mingkui Tan, Lingqiao Liu, and Ian Reid.
\newblock Towards effective low-bitwidth convolutional neural networks.
\newblock In \emph{Proceedings of the IEEE conference on computer vision and pattern recognition}, pages 7920--7928, 2018.

\end{thebibliography}
}

\end{document}